\documentclass[runningheads]{llncs}

\usepackage{eccv}

\usepackage{eccvabbrv}

\usepackage{graphicx}
\usepackage{booktabs}

\usepackage[accsupp]{axessibility}  

\usepackage{enumitem}
\newcommand{\methodname}{FlexAM\xspace}
\newcommand{\motion}{motion video\xspace}

\usepackage{hyperref}

\usepackage{orcidlink}

\begin{document}

\title{FlexAM: Flexible Appearance-Motion Decomposition for Versatile Video \\ Generation Control} 

\titlerunning{FlexAM}





\author{
Mingzhi Sheng\orcidlink{0009-0005-9062-2464}$^{1,*}$ \quad
Zekai Gu\orcidlink{0000-0002-9739-1209}$^{2,*}$ \quad
Peng Li\orcidlink{0000-0003-1066-6642}$^{2}$ \quad
Cheng Lin\orcidlink{0000-0002-3335-6623}$^{3}$ \\
Hao-Xiang Guo\orcidlink{0009-0009-0002-5252}$^{4}$ \quad
Ying-Cong Chen\orcidlink{0000-0002-9565-8205}$^{1,2,\dagger}$ \quad
Yuan Liu\orcidlink{0000-0003-2933-5667}$^{2,\dagger}$
}
\authorrunning{M.~Sheng et al.}
\institute{
$^{1}$The Hong Kong University of Science and Technology (Guangzhou), Guangzhou, China\\
$^{2}$The Hong Kong University of Science and Technology, Hong Kong SAR, China\\
$^{3}$Macau University of Science and Technology, Macau SAR, China\\
$^{4}$Tsinghua University, Beijing, China\\
\email{
msheng758@connect.hkust-gz.edu.cn,
\{zekai.gu,plibp\}@connect.ust.hk,
chenglin@must.edu.mo,
guohaoxiangxiang@gmail.com,
\{yingcongchen,yuanly\}@ust.hk
}
}

\maketitle

\begingroup
\renewcommand{\thefootnote}{}
\footnotetext{$^{*}$Equal contribution. $^{\dagger}$Corresponding authors.}
\endgroup

\begin{abstract}
  Effective and generalizable control in video generation remains a significant challenge. While many methods rely on ambiguous or task-specific signals, we argue that a fundamental disentanglement of "appearance" and "motion" provides a more robust and scalable pathway. We propose \methodname, a unified framework built upon a novel 3D control signal. This signal represents video dynamics as a point cloud, introducing three key enhancements: multi-frequency positional encoding to distinguish fine-grained motion, depth-aware encoding, and a flexible control signal for balancing precision and generalization. This representation allows \methodname to effectively disentangle appearance and motion, enabling a wide range of tasks including I2V/V2V editing, camera control, and spatial object editing. Extensive experiments demonstrate that \methodname achieves superior performance across all evaluated tasks. Codes
  are available at \url{https://github.com/IGL-HKUST/FlexAM}.
  \keywords{video generation \and video editing \and generative rendering}
\end{abstract}

\section{Introduction}
\label{sec:intro}

\begin{figure}[htbp]
  \centering
  \includegraphics[width=\linewidth]{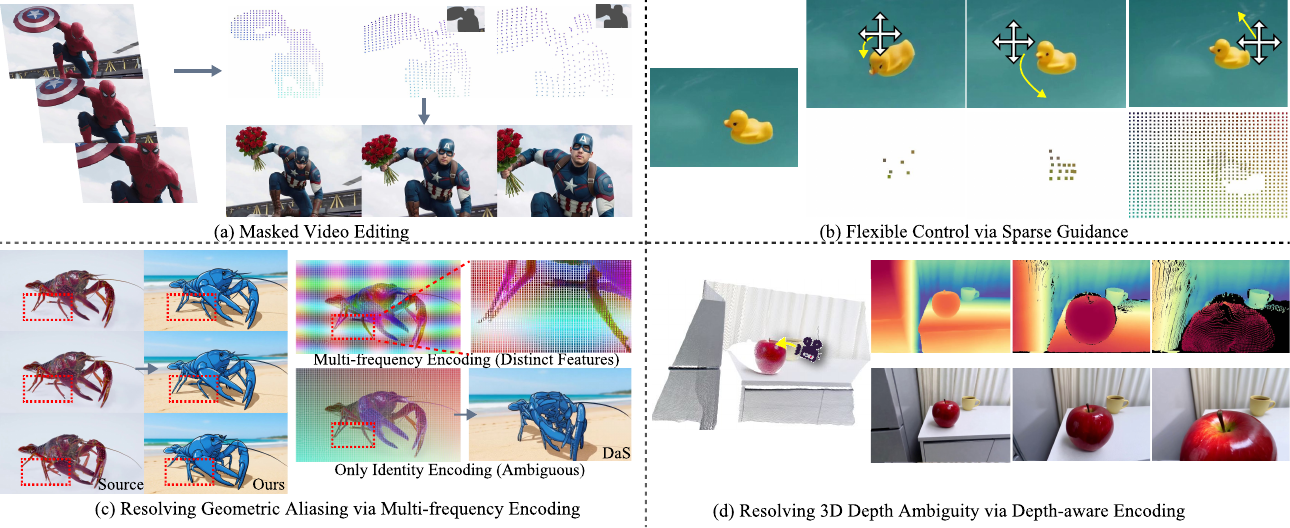}
    \caption{\textbf{\methodname} treats controllable video generation as a fundamental disentanglement of appearance and motion. By introducing a multi-attribute 3D point cloud as a robust motion representation, our framework enables: \textbf{(a)} masked video editing that transfers appearance while preserving motion; \textbf{(b)} flexible control balancing precision and generalization via varying point densities; \textbf{(c)} multi-frequency encoding to resolve geometric aliasing; and \textbf{(d)} depth-aware encoding to resolve 3D depth ambiguity.}
  
  \label{fig:teaser}
\end{figure}

The field of video generation has witnessed remarkable breakthroughs in recent years. Beyond the typical text-to-video~\cite{opensoraplan, cogvideox, ltx-video} and image-to-video~\cite{wan, longcat, ltx-video} paradigms, controllable video generation~\cite{TokenFlow, MagicEdit, vace, humo, VideoP2P, ling2024motionclonetrainingfreemotioncloning, ma2025followyourmotionvideomotiontransfer, CCEdit} has emerged as a significant area of research, spawning a multitude of downstream tasks. These include, for instance, reference-image-based~\cite{phantom, magref, hunyuancustom, humo} and reference-audio-based~\cite{wan-s2v, InfiniteTalk, FantasyTalking} video synthesis. This rapid evolution underscores the dynamic and intricate nature of the field.

To enhance controllability, researchers have explored decomposing videos into various elemental signals. Prior works have often focused on specific tasks by isolating particular signals, such as skeletons~\cite{wan-animate, InfiniteTalk}, optical flow~\cite{TokenFlow, goflow}, depth~\cite{CCEdit, MagicEdit}, camera motion~\cite{cinemaster, ReCamMaster, Uni3C, MotionCtrl, CameraCtrl}, and object trajectories~\cite{das, motionprompting, DragAnything, Tora, MagicMotion}. While some studies have attempted to combine these disparate modalities for compositional control~\cite{vace, wan-animate, VideoP2P, EditVerse, ma2025followyourmotionvideomotiontransfer}, we argue that such approaches are often inefficient. They typically require bespoke training data and model designs for each new control signal, which complicates the pipeline and limits scalability. For instance, while VACE employs a unified architecture, it still requires distinct types of control inputs tailored to specific tasks, such as depth maps for structural control and human skeletons for pose transfer. This necessity for task-specific data preprocessing to extract various modalities—and the subsequent requirement for multi-condition training—complicates the data pipeline and limits the model's ability to generalize to unseen control types.

A more fundamental approach is to decompose video into ``appearance'' and ``motion.'' We posit that this disentanglement is essential for robust controllable generation, as ``motion'' is a generalizable concept that can encapsulate human pose, camera movement, and object dynamics. Recent pioneering works have adopted this philosophy; for instance, Tora~\cite{Tora} models motion via 2D trajectories, while MagicMotion~\cite{MagicMotion} utilizes object bounding boxes. More recently, Diffusion as Shader (DaS)~\cite{das} introduced a 3D point cloud as a more versatile motion signal.

However, current appearance-motion decomposition methods face four critical limitations: 
(1) \textbf{First-frame Dependency}: Reliance on initial-frame conditioning prevents these models from describing or maintaining the appearance of regions that emerge later in the sequence. 
(2) \textbf{Inflexible Control}: Existing methods typically operate at a fixed granularity, forcing a trade-off between dense structural precision and sparse generative flexibility (Figure~\ref{fig:teaser}b). Most frameworks fail to adaptively bridge these two control regimes.
(3) \textbf{Geometric Aliasing}: Existing motion representations cannot distinguish between the motions of closely adjacent elements, often leading to identity confusion and failure in scenarios like Figure~\ref{fig:teaser}c. 
(4) \textbf{Depth Ambiguity}: The lack of explicit depth information hinders accurate reasoning about spatial occlusions and object proximity relative to the camera, resulting in physically implausible dynamics.

To address these challenges, we propose \methodname{}, a unified framework built upon a novel appearance-motion decomposition. For appearance representation, \methodname{} moves beyond first-frame conditioning by accepting arbitrarily masked video (or individual frames) as the appearance condition. This allows it to seamlessly execute tasks including image-to-video synthesis, masked video-to-video editing, and arbitrary combinations thereof. 

For motion representation, we construct a 3D-aware ``motion video'' composed of a set of trajectories, where the initial XYZ coordinates of the points define the trajectory colors to ensure temporal consistency. This motion video incorporates three core design elements: (1) As shown in Figure~\ref{fig:teaser}c, to achieve high precision and distinguish adjacent elements, the pixel color is determined by the point's 3D coordinates in the first frame's camera system, encoded using multi-frequency positional encoding across multiple layers. (2) To ensure the signal is depth-aware, we introduce additional explicit layers that represent the depth variations of these points over time, enabling the model to better reason about 3D structure and occlusion. 
(3) \methodname{} supports varying point densities to represent motion, enabling it to flexibly balance precision and generalization to meet diverse task requirements. This is achieved through a training strategy that involves varying degrees of downsampling on the 3D control signal. In summary, our contributions are as follows:
\begin{enumerate}[nosep]
    \item We propose a comprehensive and general-purpose 3D control signal. By abstracting video motion into a colored dynamic point cloud and using precise multi-frequency positional encodings to distinguish adjacent elements, it explicitly represents the 3D spatio-temporal motion of elements. This enables our model to precisely control the motion and trajectories of elements during generation and editing, facilitating tasks such as camera control and object manipulation.
    
    \item Beyond single-frame constraints, our method leverages masked video to provide a unified appearance condition for I2V and V2V tasks. By effectively disentangling appearance and motion, we enable flexible, independent editing of these components for any part of the input video or image.
    
    \item We employ a training strategy that involves varying degrees of downsampling on the 3D control signal. This allows the model to learn to balance precision and generalization, keeping flexible and adapting to the demands of different tasks.
\end{enumerate}
\noindent We conduct extensive experiments comparing \methodname{} with baseline methods on tasks including controllable video generation and editing, camera control, and spatial object editing. The results demonstrate that \methodname{} surpasses all baselines and achieves state-of-the-art (SOTA) performance across all evaluated tasks.

\section{Related Work}
\label{sec:related_work}

\subsection{Controllable Video Generation}

Recent advancements in video diffusion models have achieved high-fidelity, temporally coherent video generation~\cite{wan, cogvideox, opensoraplan}, spurring research into controllable video synthesis. Early efforts focused on specific modalities\cite{Tora, TokenFlow, Uni3C, VideoP2P, MotionCtrl, CCEdit}, such as using poses or skeletons for human-centric animation~\cite{humo, EchoMimicV2, wan-animate}, or audio for talking heads and sound-synchronized scenes~\cite{wan-s2v, InfiniteTalk, FantasyTalking}. More recent paradigms include trajectory-based control, where users specify 2D motion paths~\cite{DragAnything, Tora, motionprompting}, and 3D-aware control, targeting cinematic camera or unified human-camera motion~\cite{cinemaster, Uni3C, ReCamMaster, MotionCtrl, CameraCtrl}.
Despite this progress, most existing methods are designed for a single type of control and often rely on 2D signals, which can be ambiguous when representing complex 3D spatial dynamics.

\subsection{Appearance Editing} 
\label{sec:related_app_editing} 
Appearance editing aims to modify visual attributes, such as object identity or style, while preserving the original motion. A key challenge in this task is effectively decoupling the appearance and motion.
Previous approaches use various strategies to achieve this. Video-P2P~\cite{VideoP2P} and follow-your-motion~\cite{ma2025followyourmotionvideomotiontransfer} adapt an image model using cross-attention control. MagicEdit~\cite{MagicEdit}, Motion Prompting~\cite{motionprompting}, and MotionCtrl~\cite{MotionCtrl} explicitly disentangle content, structure, and motion during training. However, these methods often rely on intermediate 2D representations. These 2D signals are usually lossy and lack 3D spatial context, leading to ambiguities and artifacts during complex motion. More comprehensive frameworks like VACE~\cite{vace} aim for versatility but are not truly unified, requiring users to prepare a cumbersome and diverse set of conditions, such as depth maps, skeletons, or masks.
\methodname addresses these limitations by using the \motion as an explicit representation of the underlying dynamics. This single, 3D-aware signal avoids both the ambiguity of 2D representations and the complexity of multiple conditions. By conditioning on this unified signal, our model better preserves the target motion while propagating the new appearance to the masked regions.

\subsection{Camera Control}
\label{sec:related_camera_control}
Precise camera control is essential for cinematic storytelling and requires a strong understanding of 3D scene geometry. Recent works, however, often specialize in distinct sub-tasks. For instance, CineMaster~\cite{cinemaster} and CameraCtrl~\cite{CameraCtrl} present 3D-aware frameworks for I2V/T2V generation with controllable trajectories, while ReCamMaster~\cite{ReCamMaster} focuses on V2V re-rendering. This specialization means I2V-focused models cannot perform V2V re-cinematography, and vice versa.
Other methods have explored different avenues for 3D-aware control. DaS~\cite{das} aimed for versatile control, including camera manipulation, using a 3D tracking video.
However, these methods suffer from limitations, including early 3D-aware models, which implicitly infer 3D structure, leading to potential depth ambiguity. The 3D tracking signal lacked explicit, time-varying depth information, which resulted in instabilities during precise camera control. This highlights the need for a framework that not only unifies I2V and V2V camera control but also relies on a 3D signal that is robust against geometric ambiguity.

\subsection{Spatial Object Editing}
\label{sec:related_object_editing}
Spatial object editing involves manipulating the 3D position, rotation, and trajectory of specific objects within a scene. Some 3D-aware methods approach this task by focusing on explicit geometry. GeoDiffuser~\cite{sajnani2025geodiffusergeometrybasedimageediting} enables object manipulation by directly editing the scene's depth maps. While this provides explicit 3D control, such an approach can lack flexibility. DaS~\cite{das} introduced 3D tracking videos as a promising 3D-aware control signal. However, this earlier 3D representation lacked sufficient depth detail to resolve complex 3D relationships, leading to spatial inconsistencies. \methodname{} addresses this limitation by proposing a more comprehensive and flexible 3D control signal, encoding richer spatial and depth information to achieve fine-grained, spatially-consistent object manipulation.

\begin{figure*}[!b]
    \centering
    \includegraphics[width=1\linewidth]{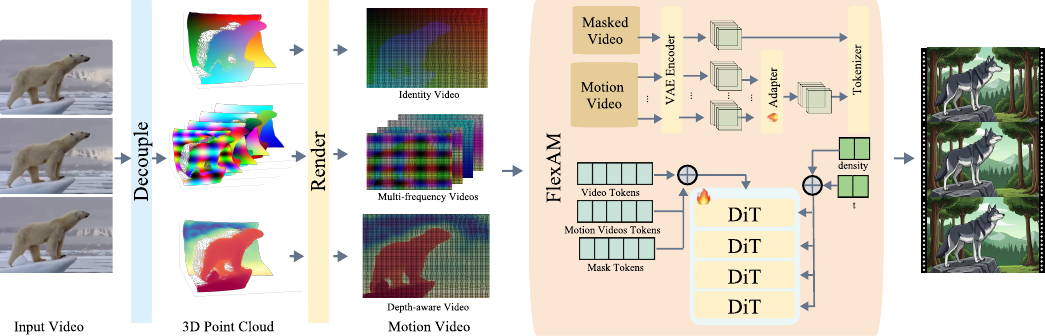}
    \caption{The \methodname{} pipeline. FlexAM decomposes controllable video generation into appearance and motion control. Given an input video, we construct a masked video as the appearance condition and render the 3D point cloud into an 18-channel motion video as the motion condition. These decoupled signals are encoded by the VAE, fused through the Adapter and tokenizer, and injected into the DiT backbone to generate the edited video while preserving the source motion dynamics.}
    \label{fig:pipeline}
\end{figure*}

\section{Method}
\label{sec:method}

Our approach, \methodname{}, is built upon the key insight of disentangling video into two fundamental components: appearance and motion. This separation is pivotal, as it permits the independent editing of either part. \methodname{} can then utilize these edited representations as control signals to the generative model, enabling targeted adjustments to the video's appearance or motion while keeping another part of the video consistent. We first detail our representations for appearance (Sec.~\ref{sec:appearance_rep}) and our novel \motion{} (Sec.~\ref{sec:motion_rep}), followed by the architecture that integrates them (Sec.~\ref{sec:architecture_training}).

\subsection{Appearance Control Signal Representation}
\label{sec:appearance_rep}
We define the appearance broadly to accommodate diverse generation and editing tasks. Unlike methods conditioned solely on a single first frame, \methodname{} accepts a partially masked video $\mathbf{V}_{\text{masked}} \in \mathbb{R}^{T \times H \times W \times 3}$ alongside a corresponding binary edit mask $\mathbf{f}_{\text{mask}}$ as the appearance condition. In $\mathbf{V}_{\text{masked}}$, unedited regions retain their original pixel values, while regions targeted for editing are filled with a gray value of 127. The binary mask $\mathbf{f}_{\text{mask}} \in \mathbb{R}^{T \times H \times W \times 1}$ explicitly delineates these targeted areas, where a value of $1$ exactly aligns with the 127-filled regions in $\mathbf{V}_{\text{masked}}$.  $\mathbf{f}_{\text{mask}}$ serves as an additional condition to apply distinct denoising timesteps for edited versus preserved regions, ensuring accurate modification while strictly maintaining unedited contexts. This flexible representation allows the model to naturally handle tasks ranging from image-to-video, where all frames except the first are masked, to local video-to-video editing, all within a unified framework.

\subsection{Motion Control Signal Representation}
\label{sec:motion_rep}

Inspired by the 3D point cloud representation for video dynamics~\cite{das}, we propose a multi-attribute motion representation. Unlike prior single-dimensional tracking signals, our points are endowed with high-dimensional attributes designed for precise, depth-aware, and flexible control. As shown in Figure~\ref{fig:pipeline}, this attributed point cloud is rendered into an 18-channel temporal signal, which we term the \motion{}, to serve as our unified motion control interface. The 18 channels are organized as six RGB-like streams: one identity stream, four multi-frequency streams, and one depth-aware stream.

Specifically, the \motion{} describes the dynamics of a set of 3D points $\{\mathbf{p}_i(t) \in \mathbb{R}^3\}$, where $t$ is the frame index. Each point $\mathbf{p}_i(t) = (x_i(t), y_i(t), z_i(t))$ corresponds to a specific spatial region. To create the final control signal, these points are projected onto the 2D screen space. At each projected location $(u_i(t), v_i(t))$, we populate attribute vector $\mathbf{A}_i(t)$. All locations without projected points are set to zero. This attribute vector $\mathbf{A}_i(t)$ is composed of three components, strategically designed to be precise, depth-aware, and flexible.

\noindent\textbf{Precise Positional Encoding} To precisely distinguish adjacent points and capture fine-grained spatial relationships, we encode the initial 3D coordinates. This includes an identity positional encoding $\mathbf{f}_{\text{identity}} \in \mathbb{R}^3$, which stores the normalized initial coordinates $(x'_i(0), y'_i(0), z'_i(0))$, where $z'$ represents inverse depth. Additionally, we compute a multi-frequency positional encoding $\mathbf{f}_{\text{freq}} \in \mathbb{R}^{12}$ by concatenating four RGB-like streams, where each stream applies $\gamma_l(v)=\cos(2^l\pi v)$, $l\in\{0,1,2,3\}$, to the three normalized initial coordinates.

\noindent\textbf{Depth-aware Representation} To explicitly encode 3D structure and resolve depth ambiguity, we include a depth-aware encoding $\mathbf{f}_{\text{depth}} \in \mathbb{R}^3$. This component represents the current normalized depth $z''_{i}(t)$ at time $t$, which is mapped to an RGB vector using the Spectral colormap $\mathcal{S}(\cdot)$. This offers richer gradients than grayscale, enhancing sensitivity to fine-grained depth variations and accelerating convergence.

\noindent\textbf{Control Flexibility and Masked Editing} During 2D rendering, points projected onto unedited regions ($\mathbf{f}_{\text{mask}} = 0$) are zeroed out, confining the motion signal to the target areas. For the edited regions, the full attribute vector $\mathbf{A}_i(t)$ at the projected location $(u_i(t), v_i(t))$ is the concatenation of these components:
\begin{equation}
    \mathbf{A}_i(t) = \left( \mathbf{f}_{\text{identity}}^{(i)}, \mathbf{f}_{\text{freq}}^{(i)}, \mathbf{f}_{\text{depth}}^{(i,t)} \right) \in \mathbb{R}^{18},
    \label{eq:attribute_vector}
\end{equation}
where
\begin{alignat}{2}
    &\mathbf{f}_{\text{identity}}^{(i)} &&= (x'_i(0), y'_i(0), z'_i(0)) \in \mathbb{R}^{3}, \label{eq:f_identity} \\
    &\mathbf{f}_{\text{freq}}^{(i)}     &&= \operatorname{Concat}_{l=0}^{3}
    \left(
    \gamma_l(x'_i(0)), \gamma_l(y'_i(0)), \gamma_l(z'_i(0))
    \right) \in \mathbb{R}^{12}, \label{eq:f_freq} \\
    &\mathbf{f}_{\text{depth}}^{(i,t)}  &&= \mathcal{S}(z''_{i}(t)) \in \mathbb{R}^{3}. \label{eq:f_depth}
\end{alignat}

\noindent\textbf{Sparse-to-Dense Flexibility.}
A key feature of the \motion{} is its density flexibility. The signal can represent motion densely or sparsely, determined by the number of points $\{\mathbf{p}_i(t)\}$ used. This allows \methodname{} flexibly to balance precision and generalization. As detailed in Sec.~\ref{sec:architecture_training}, we train the model to handle this variability by stochastically down-sampling the point clouds during training.

\subsection{Architecture and Training Integration}
\label{sec:architecture_training}
To inject our decoupled signals into a pre-trained latent video diffusion model (e.g., Wan2.2Fun 5B Control), we encode both appearance and motion conditions into the backbone's latent space. Specifically, the appearance condition $\mathbf{V}_{\text{masked}}$, together with its binary mask $\mathbf{f}_{\text{mask}}$, is encoded into appearance latents, while the 18-channel motion video is first partitioned into six RGB-like streams and then encoded independently using the same pre-trained VAE. Since the diffusion backbone operates in the VAE latent space, this shared encoding scheme makes the motion latents directly compatible with the denoising network, without introducing an additional task-specific motion encoder. A lightweight CNN-based Adapter then fuses the motion latents into a unified motion control signal $\mathbf{C}$. To support flexible sparse-to-dense control, we introduce a density embedding for the point density $d$. Following the design of the diffusion timestep embedding, $d$ is mapped to an embedding vector and injected into the denoising network, explicitly indicating the spatial density of the motion control signal. Detailed training configurations are provided in the supplementary material.

\subsection{Controllable Video Generation and Editing}
\label{sec:task}

In this section, we elaborate on how \methodname achieves controllable video generation and video editing.

\noindent\textbf{Appearance Editing.}
\label{sec:app_video_editing}
Appearance editing aims to modify the visual style or specific regions of an input video (e.g., style transfer or character replacement) while preserving its original motion dynamics. 
We achieve this by leveraging \methodname's decoupled representation: extracting the source dynamics as the motion interface, and using the masked video to guide the appearance generation. 
Specifically, we first detect 3D points and their trajectories using DELTA \cite{delta} from the source video to generate a \motion. Next, we construct the appearance condition by masking the editable regions (filling them with gray) and prepending a repainted first frame to the sequence. Finally, \methodname integrates these inputs to transfer the target appearance to the masked regions according to the preserved dynamics, generating the target video.

\noindent\textbf{Camera Control.}
\label{sec:app_camera_control}
This task aims to synthesize videos from static images or existing videos that follow a user-specified camera trajectory. 
Our core idea is to construct a 3D or 4D dynamic point cloud of the scene and re-project these points from the novel camera views to form the \motion.

For the image-to-video generation task with a specific camera trajectory, we first estimate the depth map of the input image, which is then encoded using $\mathbf{A}_i(t)$.
These points are then projected onto the given camera trajectory to construct a \motion, allowing \methodname to control the camera motion precisely.

For the video-to-video camera control task, we first detect 3D points and trajectories and estimate the camera parameters and trajectory of the input video.
Then we combine them to build a 4D dynamic point cloud.
Subsequently, these points are projected onto the new camera trajectory to construct a new \motion for \methodname to achieve high-precision camera re-cinematography.

\noindent\textbf{Spatial Object Editing.}
\label{sec:app_object_editing}
Spatial object editing involves manipulating the 3D position, rotation, and scale of specific objects within a scene while maintaining geometric plausibility. 
Our framework accomplishes this by isolating the 3D point cloud of the target object, applying explicit geometric transformations in 3D space, and rendering the updated scene geometry into a new \motion. 
Given an input image, we first estimate its depth map and segment the target object. We then apply the desired geometric manipulations directly to the point cloud of this isolated object. Finally, utilizing the depth-aware encoding attributes to avoid depth ambiguity, we project the manipulated point cloud back to the image plane to construct the \motion for the final video generation.

\section{Experiment}
We conduct experiments across three tasks, appearance editing, camera control, and spatial object editing, to demonstrate the versatility of \methodname in video editing and controlling video generation. 

\subsection{Appearance Editing}

\begin{figure}[!b]
  \centering
  \includegraphics[width=1\linewidth]{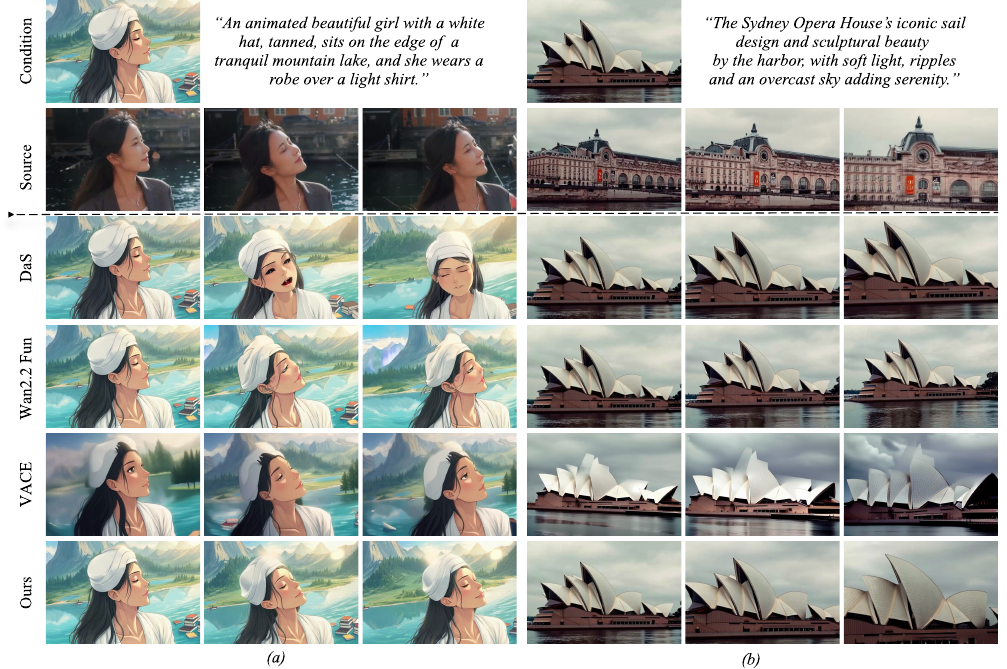} 
  \caption{Qualitative comparison on motion transfer between our method, DaS, Wan2.2 Fun, and VACE. We compare the results of different methods in transferring the human motion from the Source to a new appearance. Compared to the baseline, our method accurately transfers the motion.}
  \label{fig:motiontransfer} 
\end{figure}

We primarily compared \methodname and the baseline on two subtasks: motion transfer and partial editing. For a fair comparison, all models use the same reference images generated by Qwen Image Edit~\cite{qwenimgedit} as the unified appearance guidance.

\noindent\textbf{Motion Transfer.} \noindent\textit{Baselines}. We compare \methodname with representative motion transfer methods, including DaS~\cite{das}, VACE~\cite{vace}, and Wan2.2Fun (5B) Control~\cite{VideoXFun2024}. Specifically, VACE and Wan2.2Fun are conditioned on depth maps, while DaS utilizes 3D tracking videos. In contrast, our method is conditioned on the proposed \motion{}, which provides a more comprehensive 3D-aware representation.

\noindent\textit{Metrics.} We evaluate text-video alignment (Tex-Ali) and temporal consistency (Tem-Con) using CLIP~\cite{radford2021learningtransferablevisualmodels, hessel2022clipscorereferencefreeevaluationmetric}. Additionally, we employ VBench++~\cite{huang2025vbench++} to comprehensively assess subject/background consistency (\textit{i2v\_sub}, \textit{i2v\_bg}), motion smoothness (\textit{mot\_sm}), and aesthetic/imaging quality (\textit{aes\_ql}, \textit{img\_ql}).

\begin{table}[htbp]
    \caption{Quantitative comparison of Motion Transfer. \methodname~achieves the highest text-video alignment, temporal consistency, and visual aesthetics while maintaining high subject and background fidelity.}
    \centering
    \scriptsize 
    \setlength{\tabcolsep}{3.5pt} 
    \renewcommand{\arraystretch}{0.85} 
    \begin{tabular}{lcc|ccccc}
        \toprule
        & \multicolumn{2}{c|}{\textbf{CLIP Metrics}} & \multicolumn{5}{c}{\textbf{VBench++ Metrics}} \\
        \cmidrule(lr){2-3} \cmidrule(lr){4-8}
        Method & Tex-Ali $\uparrow$ & Tem-Con $\uparrow$ & i2v\_sub $\uparrow$ & i2v\_bg $\uparrow$ & mot\_sm $\uparrow$ & aes\_ql $\uparrow$ & img\_ql $\uparrow$ \\
        \midrule
        DaS~\cite{das} & 32.14 & 0.968 & 0.98 & 0.98 & 0.98 & 0.60 & \textbf{0.68} \\
        Wan2.2 Fun~\cite{VideoXFun2024} & 32.39 & 0.971 & 0.98 & \textbf{0.99} & 0.98 & 0.61 & \textbf{0.68} \\
        VACE~\cite{vace} & 32.38 & 0.970 & 0.86 & 0.86 & \textbf{0.99} & 0.57 & 0.63 \\
        \textbf{\methodname~(Ours)} & \textbf{32.55} & \textbf{0.976} & \textbf{0.99} & \textbf{0.99} & \textbf{0.99} & \textbf{0.64} & \textbf{0.68} \\
        \bottomrule
    \end{tabular}
    \label{tab:motion_transfer_performance}
\end{table}

\noindent\textit{Results.} As shown in Table~\ref{tab:motion_transfer_performance}, \methodname~exhibits overall superior performance. It significantly outperforms VACE in subject and background fidelity (0.99 vs. 0.86), alongside a notable improvement in aesthetic scores (\textit{aes\_ql}: 0.64).  Qualitatively (Figure~\ref{fig:motiontransfer}), VACE repaints the reference image and loses character identity, Wan2.2Fun distorts facial features due to over-alignment with depth, and DaS fails to reconstruct the pose. In contrast, \methodname~accurately transfers motion with superior visual harmony and temporal coherence.

\begin{figure*}[!b]
    \centering
    \includegraphics[width=1\linewidth]{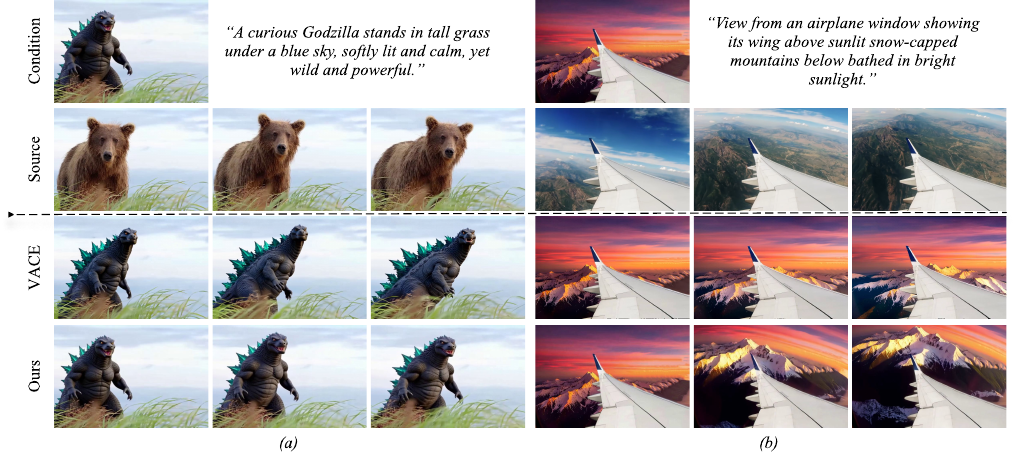}
    \caption{\textbf{Qualitative comparison on foreground and background editing.} We transfer the motion from the source videos while replacing the foreground/ background appearance using the reference prompt/image. (a) Replace bear with Godzilla; compared to VACE, our method better follows the reference poses and preserves identity and color details. (b) Airplane wing over mountains at sunset; While VACE maintains foreground consistency but loses background motion, our method integrates the input video’s background motion with the new appearance, preserving coherent dynamics.}
    \label{fig:partialedit}
\end{figure*}

\begin{figure*}[htbp]
    \centering    \includegraphics[width=1\linewidth]{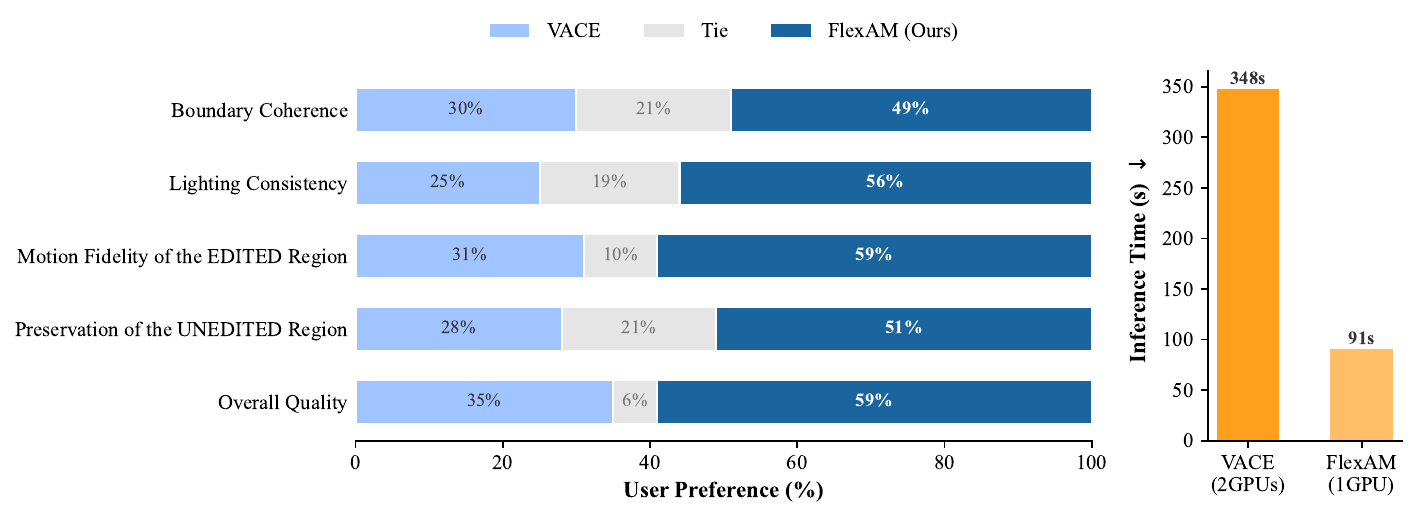}
    \caption{\textbf{Quantitative comparison on partial editing.} We compare the user preference (\%) and inference time between our method and VACE. (Left) \methodname is consistently preferred across all dimensions. (Right) Compared to VACE, \methodname achieves a significant speedup, demonstrating superior efficiency.}
    \label{fig:user_study_partial}
\end{figure*}

\noindent\textbf{Partial Editing.}
This task targets modifying the appearance of specific regions while preserving their original motion trajectories, a capability essential for applications such as virtual try-on and cinematic customization.

\noindent\textit{Baselines.} We compare \methodname~with VACE on both foreground and background editing. Specifically, we use SAM2~\cite{SAM2} to mask the foreground or background of the input videos. The models are expected to accurately transfer the appearance from the reference images to the masked regions, seamlessly integrate it with the original motion, and faithfully preserve the unmasked areas.

\noindent\textit{Results.} As shown in Figure~\ref{fig:partialedit}(a), during foreground editing, VACE frequently redraws the first-frame appearance and exhibits temporal drift at limb boundaries. In contrast, \methodname{} better follows the reference rhythm while maintaining more stable identity and local details. For background editing (Figure~\ref{fig:partialedit}(b)), VACE fails to preserve background motion, while \methodname{} integrates the reference background appearance with the source background dynamics and maintains foreground consistency. We further conduct a randomized A/B user study with 27 participants on 8 video pairs. As shown in Figure~\ref{fig:user_study_partial} (Left), \methodname{} is consistently preferred, particularly in lighting consistency (56\% vs. 25\%) and preservation of the unedited region (51\% vs. 28\%).  Efficiency evaluation in Figure~\ref{fig:user_study_partial} (Right) further highlights our advantages: despite a 14B baseline, our 5B architecture requires only a single GPU and significantly reduces inference time from 348s to 91s. Additional quantitative results using VBench++~\cite{huang2025vbench++} and VE-Bench~\cite{sun2024bench} are provided in the supplementary material. 

\subsection{Camera Control}
We evaluate camera control to examine whether the proposed general-purpose motion representation can achieve competitive geometric precision against task-specific camera-control models. We conduct quantitative evaluation on image-to-video (I2V) camera control for trajectory accuracy, and further provide qualitative results on the more challenging video-to-video (V2V) camera re-rendering setting.

\noindent\textit{Baselines.}
We compare \methodname{} with three representative camera-control methods: DaS~\cite{das}, Wan2.2 Fun Control Camera~\cite{VideoXFun2024}, and ReCamMaster~\cite{ReCamMaster}. All methods are given the same input frame or video and the same target camera trajectory whenever applicable. Since Wan2.2 Fun Control Camera only supports image-to-video generation, it is excluded from the qualitative V2V comparison.

\noindent\textit{Metrics.}
To evaluate camera-control accuracy, we measure the discrepancy between the target camera trajectory and the trajectory estimated from the generated video. Specifically, for each generated video, we use Pi3~\cite{pi3} to reconstruct the relative camera pose of each frame with respect to the first frame. We then extract the normalized rotation quaternion and translation vector, and compute the rotation and translation errors as:
\begin{equation}
\mathrm{RotErr}=\arccos\!\Big(\frac{1}{T-1}\sum_{i=2}^{T}
\langle \mathbf{q}^{i}_{\mathrm{gen}}, \mathbf{q}^{i}_{\mathrm{gt}}\rangle\Big), \quad
\mathrm{TransErr}=\arccos\!\Big(\frac{1}{T-1}\sum_{i=2}^{T}
\langle \mathbf{t}^{i}_{\mathrm{gen}}, \mathbf{t}^{i}_{\mathrm{gt}}\rangle\Big).
\label{eq:camera_error}
\end{equation}
All methods are evaluated using the same Pi3-based pose estimation protocol and compared against the same ground-truth trajectories.

\begin{figure}[!t]
    \centering
    \includegraphics[width=\linewidth]{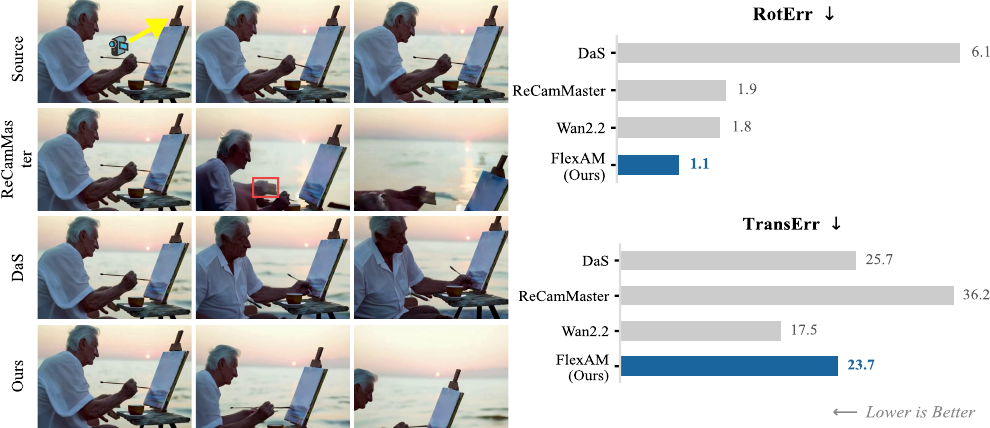}
    \caption{\textbf{Camera control comparison.}
    (Left) Qualitative V2V camera re-rendering with a prescribed ``pan up-right'' trajectory.
    (Right) Quantitative I2V camera-control evaluation on RealEstate10K.}
    \label{fig:cameractrl}
\end{figure}

\noindent\textit{Results.}
For I2V task, we evaluate all methods on 100 randomly sampled sequences from RealEstate10K~\cite{zhou2018stereo}. Each method receives the initial frame and the ground-truth camera trajectory as input, and we compare the estimated camera poses of the generated videos against the ground-truth trajectory. As shown in Figure~\ref{fig:cameractrl} (Right), \methodname{} achieves the lowest rotation error among all methods. For translation error, the task-specific Wan2.2 Fun Control Camera obtains a slightly lower score due to its explicit pose-conditioned supervision, while \methodname{} remains competitive without relying on specialized pose-control modules.

For V2V task, we perform a qualitative comparison using an internet video with a prescribed ``pan up-right'' trajectory. As shown in Figure~\ref{fig:cameractrl} (Left), ReCamMaster~\cite{ReCamMaster} produces rendering artifacts and trajectory deviations, while DaS~\cite{das} struggles to align the generated video with the target camera motion. In contrast, \methodname{} better preserves the source content while following the prescribed camera trajectory, demonstrating its robustness in video re-rendering.

\subsection{Spatial Object Editing}

\begin{figure}[!t]
    \centering
    \begin{minipage}[c]{0.61\textwidth}
        \centering
        \includegraphics[width=\linewidth]{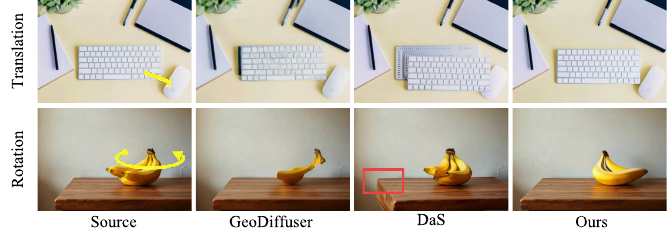}
    \end{minipage}
    \hfill
    \begin{minipage}[c]{0.37\textwidth}
        \centering
        \resizebox{\linewidth}{!}{%
            \renewcommand{\arraystretch}{1.2} 
            \begin{tabular}{lc}
                \toprule
                Method & \textbf{CLIP Score} $\uparrow$ \\
                \midrule
                GeoDiffuser~\cite{sajnani2025geodiffusergeometrybasedimageediting} & 0.9110 \\
                DaS~\cite{das} & 0.9437 \\
                \textbf{FlexAM (Ours)} & \textbf{0.9536} \\
                \bottomrule
            \end{tabular}%
        }
    \end{minipage}
    \vspace{-2mm}
    \caption{\textbf{Results for object manipulation.} \textit{Left:} Qualitative comparison of object translation (top) and rotation (bottom). \textit{Right:} Quantitative evaluation using CLIP Scores.}
    \label{fig:object_manipulation}
    \vspace{-4mm}
\end{figure}

For spatial object editing, we adopt the SAM2~\cite{SAM2} and MoGe\cite{moge} to get the object points. We compare \methodname with DaS and GeoDiffuser~\cite{sajnani2025geodiffusergeometrybasedimageediting} in two kinds of tasks: translation and rotation. Since GeoDiffuser is an image-based manipulation method, we compare its results with the final video frame generated by \methodname and DaS, and use the CLIP \cite{radford2021learningtransferablevisualmodels} to evaluate semantic alignment between the generated results and text prompts.


\noindent\textit{Results.} 
Figure~\ref{fig:object_manipulation} demonstrates that \methodname{} achieves accurate object manipulation and produces photorealistic videos with strong multiview consistency. Compared to baselines, \methodname{} exhibits better occlusion reasoning and editing capability, while also achieving the highest CLIP score. Further quantitative evaluation of spatial object editing is provided in the supplementary material.



\subsection{Ablation Study}

\begin{figure}[htbp]
    \centering
    \includegraphics[width=\linewidth]{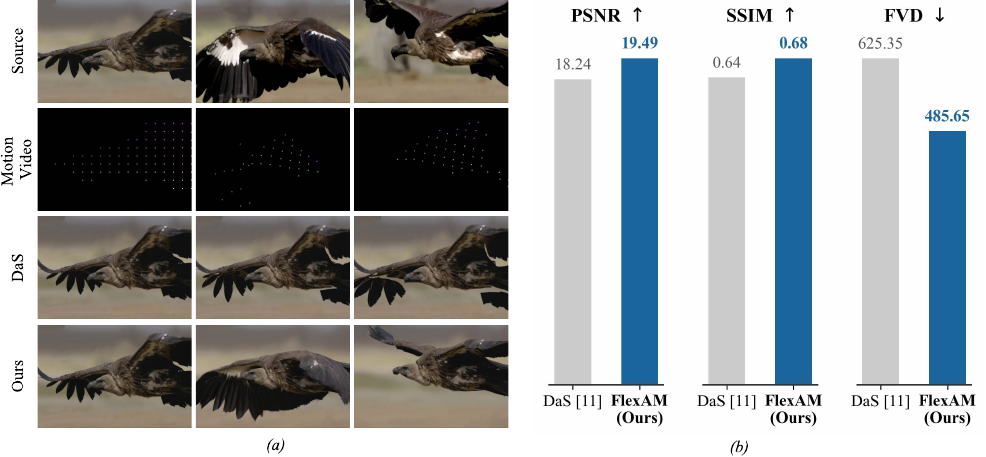}
    \caption{\textbf{Qualitative and quantitative comparison under sparse motion signals.} (a) Qualitatively, \methodname{} maintains accurate motion and appearance even under sparse control, whereas the baseline DaS~\cite{das} exhibits motion drift and shape distortion. (b) Quantitatively, \methodname{} consistently outperforms DaS across all three reconstruction metrics.}
    \label{fig:sparse}
\end{figure}

We evaluate the core components of our \motion by comparing \methodname against a baseline that utilizes only vanilla 3D Tracking Video. Specifically, we analyze the necessity of our multi-attribute design—including control flexibility, precise positional encoding, and depth-aware representation—in maintaining motion precision and spatial consistency.

\noindent\textbf{Control Flexibility.} 
We evaluate robustness to sparsity via a video reconstruction task, requiring models to recover original videos using only the first frame and extracted sparse motion signals. As presented in Figure~\ref{fig:sparse}, \methodname{}  outperforms the baseline DaS~\cite{das} in quantitative reconstruction fidelity (PSNR, SSIM, FVD). Qualitatively, our method maintains accurate motion control and source appearance even under sparse conditions, demonstrating \methodname's capability to balance flexibility and precision across varying control densities.

\begin{figure}[!t]
    \centering
    \includegraphics[width=1\linewidth]{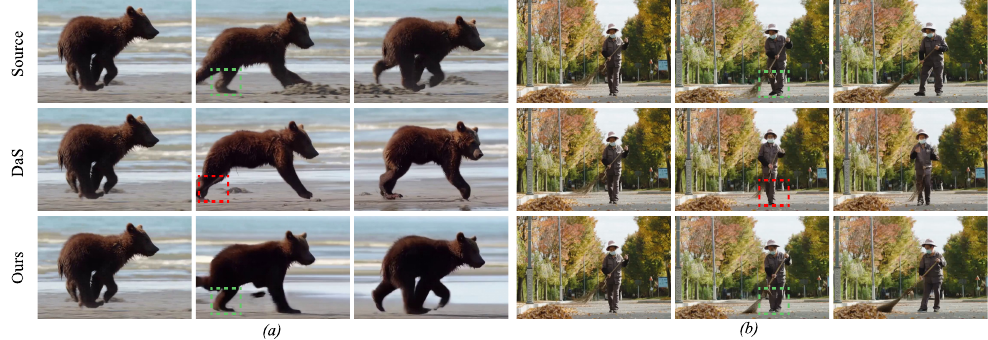}
    \caption{Qualitative comparison of precise positional encoding. \methodname adds multi-frequency encodings to keep nearby points separable and motions consistent. The areas marked with color boxes demonstrate \methodname's better ability in reconstructing dynamic details compared to the baseline.}
    \label{fig:freq_bear}
\end{figure}

\noindent\textbf{Precise Positional Encoding.} 
As shown in Figure~\ref{fig:freq_bear}, the baseline suffers from motion aliasing when screen-space trajectories overlap. This ambiguity causes the model to confuse or merge adjacent points, such as the bear's left and right feet. Our multi-frequency positional encoding resolves this, enabling the model to preserve limb identity throughout the gait cycle. This yields temporally consistent contacts and plausible kinematics, in clear contrast to the baseline which often swaps the feet.

\begin{figure}[!t]
    \centering
    \includegraphics[width=1\linewidth]{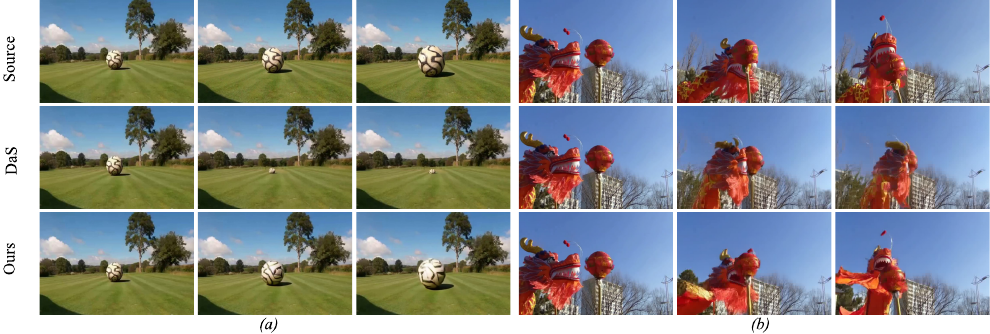}
    \caption{\textbf{Qualitative comparison of depth-aware encoding.} Without time-varying depth, scale changes can be mistaken for depth motion, leading to implausible occlusions. \methodname{} better preserves camera-object spatial relationships and produces more physically plausible dynamics.}
    \label{fig:depth}
\end{figure}

\noindent\textbf{Depth-aware Encoding.} 
Figure~\ref{fig:depth} illustrates the necessity of video depth as a conditioning factor. Encoding 3D point clouds solely based on 2D screen-space coordinates introduces ambiguity, as distinct 3D spatial movements can project to identical 2D results. Consequently, baselines relying on static depth fail to distinguish between scale variations and actual depth movements, causing physically implausible occlusions. In contrast, \methodname{} incorporates time-varying depth to accurately recover 3D trajectories and ensure physically consistent dynamics.

\begin{figure}[!t]
    \centering
    \includegraphics[width=\linewidth]{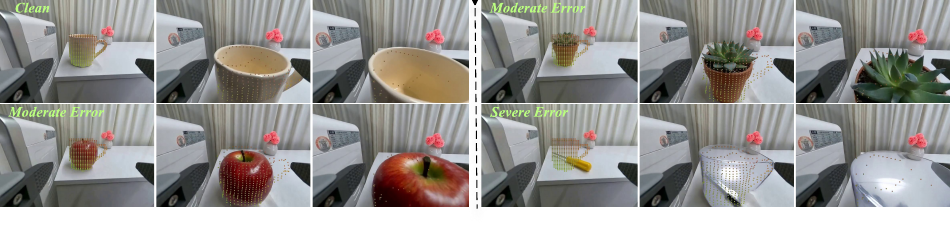}
    \caption{\textbf{Robustness and failure cases.}
    \methodname{} handles moderate geometric mismatch but degrades under severe mismatch.}
    \label{fig:robustness_failure}
\end{figure}

\section{Limitations and Conclusions}
\noindent\textbf{Limitations.} \methodname{} still faces several limitations. First, its control precision depends on the quality of the estimated 3D trajectories; inaccurate tracking, heavy occlusion or fast motion may affect fine-grained motion control. Second, although \methodname{} is robust to moderate geometric mismatch between the source motion and target appearance, severe mismatch can lead to implausible deformation, as shown in Figure~\ref{fig:robustness_failure}. Third, due to limited lighting diversity in the current training data, occasional illumination mismatches may occur during generation. Constructing more diverse lighting-aware datasets and scaling to larger video data are important directions for improving robustness.

\noindent\textbf{Conclusions.} We introduce \methodname{}, a unified framework for controllable video generation that encodes scene dynamics as a dynamic 3D point cloud augmented with multi-frequency positional encodings, depth-aware encoding and flexible control. This control representation cleanly disentangles appearance from motion, enabling a single model to handle I2V/V2V editing, camera trajectory control, and object manipulation. Extensive experiments show consistent improvements over strong baselines, indicating that FlexAM achieves high precision without sacrificing generalization.


%
%
\bibliographystyle{splncs04}
\bibliography{main}
\end{document}